# Constructing Ontology-based Cancer Treatment Decision Support System with Case-Based Reasoning

Ying Shen[1], Joël Colloc[2], Armelle Jacquet-Andrieu[3], Ziyi Guo[1], Yong Liu[4,*]

[1] School of Electronics and Computer Engineering (SECE)
Peking University Shenzhen Graduate School, 518055 Shenzhen, P.R. China
[2] CIRTAI - Université du Havre, 25 Rue Philippe Lebon, 76086, Le Havre Cedex, France
[3] MoDyCo - Université Paris Ouest (UMR CNRS 7114)
200 avenue de la République, 92000 Nanterre, France
[4] IER Business Development Center, Shenzhen, P.R. China
shenying@pkusz.edu.cn*, joel.colloc@univ-lehavre.fr,
armelle.jacquet@u-paris10.fr, guoziyi@pku.edu.cn,
13312962646@189.cn

**Abstract.** Decision support is a probabilistic and quantitative method designed for modeling problems in situations with ambiguity. Computer technology can be employed to provide clinical decision support and treatment recommendations. The problem of natural language applications is that they lack formality and the interpretation is not consistent. Conversely, ontologies can capture the intended meaning and specify modeling primitives. Disease Ontology (DO) that pertains to cancer's clinical stages and their corresponding information components is utilized to improve the reasoning ability of a decision support system (DSS). The proposed DSS uses Case-Based Reasoning (CBR) to consider disease manifestations and provides physicians with treatment solutions from similar previous cases for reference. The proposed DSS supports natural language processing (NLP) queries. The DSS obtained 84.63% accuracy in disease classification with the help of the ontology.

**Keywords:** Clinical decision support · Data mining · Ontology construction · Decision Support system · Case-Based Reasoning.

## 1 Introduction

Clinicians cannot obtain knowledge regarding clinical processes from the existing, inadequate medical databases. In current decision support systems (DSSs) [1][2], databases and computer records work together to facilitate decision-making by improving access to relevant data through defined interfaces. We focus on researching the problem of classification and diagnosis. After investigating the uncertainty about the actual situation of the study object (patient, organ, population), it is necessary to distinguish the possible symptoms and diseases from the impossible ones to determine effective measures [3].

This study proposes an ontology-based DSS to aid in cancer diagnosis and therapy. Several existing ontologies and high-quality data resources are available for use, including the Disease Ontology (DO) [4], the NCBI organismal classification ontology (NCBI Taxonomy) [5], the Human Phenotype Ontology (HPO) [6], and DrugBank [7], as well as some useful websites such as the U.S. National Library of Medicine (NLM) [8] and Wikipedia [9]. We utilize ontology triples to improve the reasoning ability of the DSS. Afterwards, Case-Based Reasoning (CBR) is employed to provide physicians with treatment solutions from similar previous cases for reference. A case study concerning the answering of clinical queries about gastric cancer (diagnosis, prognosis and treatment) is developed to illustrate the implementation of DSS with an ontology.

The main contributions of this work can be summed as followed:
- Base the medical reasoning framework which defines rules and interfaces, the combined operation of DSS, CBR and ontology can aide clinical decision-making.
- To simplify the semantic representation of medicines, existing biomedical ontology is reused in DSS.
- The comparison between DSS with or without ontology presents that our model can fully utilize the ontology information and provides a stable performance in diagnosis and disease classification.

The rest of this paper is structured as follows: Section 2 describes the related work. Section 3 introduces the details of the proposed methods; Experimental results and evaluations are presented in Section 4. Finally, we conclude the paper in section 5.

## 2   Related work

The application of DSS for the cooperation of ontology has been investigated in many works shown below. Farion et al. [10] used ontology-driven design to represent components of a CDSS. A prototype of the system was implemented for two clinical decision problems and settings (triage of acute pain in the emergency department and postoperative management of radical prostatectomy on the hospital ward). Haghighi et al. [11] have designed a knowledge acquisition tool to facilitate the creation and maintenance of a knowledge base by the domain expert and its sharing and reuse by other institutions. They used the Unified Medical Language System (UMLS) which contains the domain entities and constitutes the relations repository. Lee et al. [12] presented a novel fuzzy expert system for diabetes decision support application. The proposed fuzzy expert system can work effectively for diabetes decision support application. Jayaraman et al. [13] realized the drug side effects data representation and full spectrum inference using ontology in Intelligent Telehealth. The proposed model allows the doctors and caregivers to derive dynamic information about side effects avoiding costly errors caused by human interpretation. Ontology is used to prevent electronic health record error approach [14] etc. Besides, ontology can be used in other domain for decision supports. Ontology is adopted to control the intercrossed access for secure financial services on multimedia big data in cloud systems [15], and to classify cyber incident for cybersecurity insurance in financial industry [16]

In the development of a DSS for medical decision-making, an approach using a quantitative model analysis has increasingly gained attention. Some tools, technologies and concepts, such as decision trees and Bayes [17] and probability theories, can improve the operation of medical decision-making. CBR computes component similarities by exploring indexed knowledge. CBR is a qualitative and quantitative mixed model of experience storage and retrieval. This method is analogous to problem-solving methods that compare new cases with previously indexed cases [18]. CBR involves semantic distances developed using different approaches, including algorithms of structural similarity, statistical learning, digital approaches from neural networks, and fuzzy logic. The research on semantic distances often combines symbolic and numeric aspects [19].

## 3 Methods

### 3.1 Architecture of DSS

The architecture of DSS is presented in Fig. 1 and consists of the knowledge acquisition model, the ontology model and the interconnection model. Each part can be considered as an independent computing environment. The knowledge acquisition module provides medical data to the "Interpreter module" and the "Inference engine". The DSS repository recognizes, processes and stores medical data. The "Interpreter module" operates as an algorithm controller to perform query analysis and knowledge extraction. The "Inference engine" mainly normalizes the interrogation flow and formulates the relationships among the diagnosis, prognosis and treatment. It uses a case-based inference rule to analyze the likelihood and symptoms of complications of the current diseases so that medical care personnel can determine a prognosis or implement preventive measures.

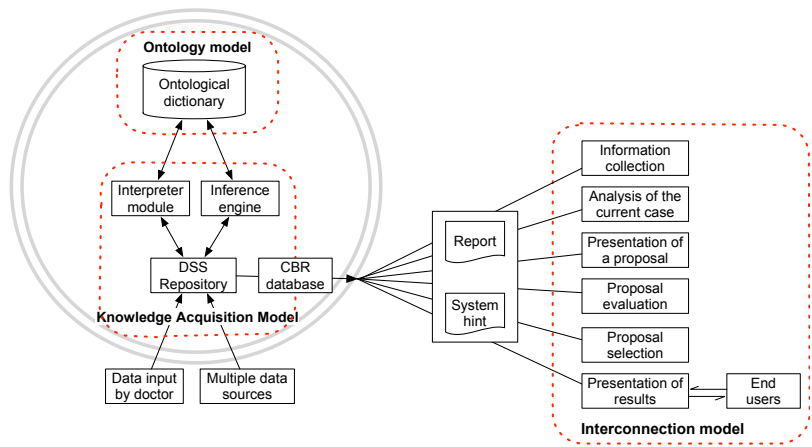

**Fig. 1.** Architecture of DSS

## 3.2 Medical Reasoning Framework

Based on our previous work [3], Fig. 2 presents the relations between diagnosis ($\Delta$), prognosis ($\Pi$), and treatment ($\Theta$) and describes a series of medical inference rules using a simulated treatment approach.

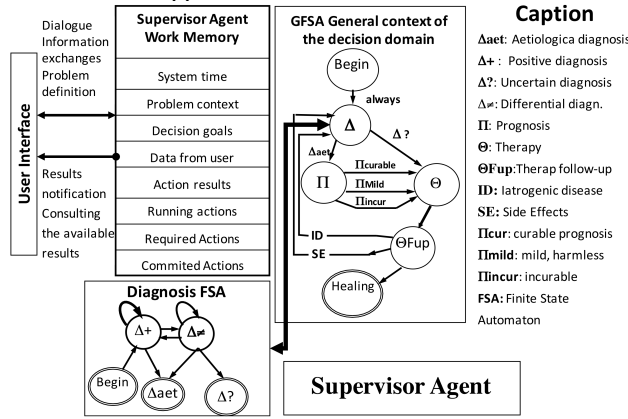

**Fig. 2.** Reasoning between diagnosis, prognosis and treatment

CBR successively stores and indexes clinical cases and knowledge in different directories ($\Delta$, $\Pi$, $\Theta$, S$\Theta$) by identifying keywords related to the problem of the case (PB), the environment (patient record) (E) and the result (R). (Fig.3)

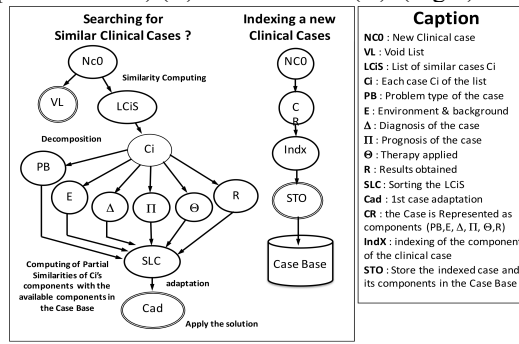

**Fig. 3.** Case-base reasoning

The supervisor monitors and triggers all necessary steps of the clinical decision process through finite state automaton (FSA) and ensures the dialogue between the computer and the end-user. It controls the management and execution of clinical tasks with predefined available models. A model is instructed to give up a task if the situation or the operating environment is amended.

In the DSS, each model uses reflexive knowledge to determine whether it should contribute to the task requested by the supervisor. If the model determines that it should contribute, then the supervisor assigns its part of the on-going actions, and the necessary models become active. Answers are extracted from the DSS repository and the CBR database and listed for query answering through a matching approach by consulting the indexed corpus.

### 3.3 Ontology Utilization

In the DO, each ontology triple is established based on an inference rule. The DO triple is extracted and utilized in this study to clarify the medical knowledge and improve the reasoning ability of the DSS. For example, the triple (radical surgery, cure, cancer) refers to the possibility that radical surgery can help to cure cancer. By means of inheritance and matching, the cancer-relevant knowledge contains the following information.

1. Symptoms of each type of cancer.
2. Classification, biological feautres and cell proliferation dynamics of cancer.
3. Cancer therapies: surgical treatment (radical surgery, palliative surgery, surgical exploration), radiotherapy, chemotherapy (systemic chemotherapy, adjuvant chemotherapy, neoadjuvant chemotherapy, special approach chemotherapy for advanced stage or disseminated tumors) and food therapy.
4. Basic therapeutic principles for tumors, including therapeutic regimens for head and neck neoplasms, thoracic neoplasms, abdominal malignant tumors, tumors of the urinary system, tumors of the female genital system, CNS tumors, malignant tumors of the hematopoietic system, soft tissue tumors, primary malignant bone tumors, and metastatic tumors. Consider breast cancer as an example:
   Phase 0 and I: Breast-conserving conservative surgery + postoperative radical radiotherapy or modified radical mastectomy.
   Early Phase II: Same as Phase 0 and I. Chemotherapy or endocrine therapy will be performed according to pathology and receptor conditions.
   Phase II: Modified radical mastectomy ± radiotherapy ± chemotherapy ± endocrine therapy.
   Phase III: Neoadjuvant chemotherapy ± radiotherapy + modified radical mastectomy (or radical resection) + postoperative radiotherapy + chemotherapy ± endocrine therapy.
   Phase IV: Mainly chemotherapy and endocrine therapy ± local radiotherapy ± local operation.
5. Diagnosis and treatment for cancer pain: Mechanism and classification of cancer pain, evaluation (conventional evaluation principle, quantitative evaluation principle, comprehensive evaluation principle and dynamic evaluation principle) of cancer pain, and therapeutic principles and methods (etiological treatment, drug analgesia therapy and non-drug therapy) for cancer pain.
6. Pharmacological action and pharmacokinetics of drugs.
7. Drug options and doses, dose intensity, relative dose intensity, dose density, course of treatment, intervals, etc.

Since new clinical knowledge cannot be applied automatically without clinical verification, a hierarchy expansion is implemented for incorporating new input data, which should be supported by clinical evidence and supervised by a knowledge engineer.

## 4 Experiments and evaluation

### 4.1 Example: Diagnosis, Prognosis and Treatment of Gastric Cancer

The output interface of the DSS is mainly used to generate query results. Fig. 4-6 illustrate the diagnosis, prognosis and treatment of gastric cancer. In Fig.4, SAT Δ concerns gastric cancer diagnosis. Its inference graph refers to the evolution of existing gastric cancer cases, corresponding to the structure of SAT Δ detailed in Fig. 2. In the inference graph, Δ0 indicates a positive diagnosis, Δ1 specifies a differential diagnosis, and Δ2 to Δ6 identify etiological diagnoses. The initial presence of gastric cancer (SO1) can lead to diagnoses Δ1 to Δ6 according to clinical signs SE1 to SE6.

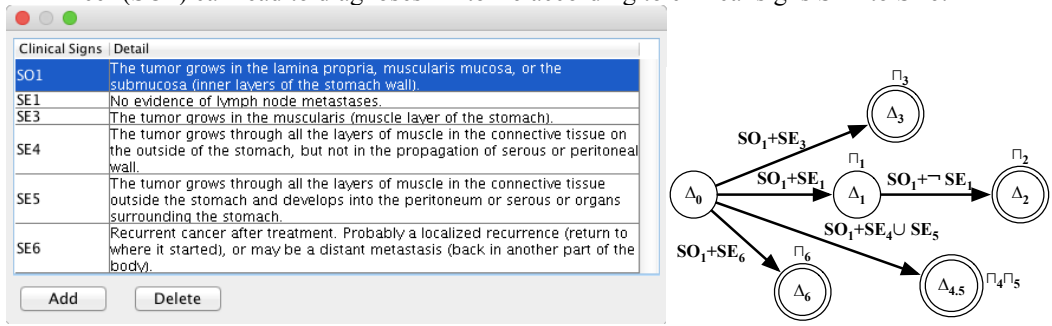

**Figure 4** Clinical signs to determine gastric cancer

SAT Θ indicates various therapeutic strategies (Θ1 to Θ6) (Fig.5) corresponding to different diagnoses and prognoses. The knowledge encapsulated in these several steps is stored in the DSS and can be extracted by CBR.

**Figure 5** Diagnoses and therapies for curing gastric cancer

Fig.6 shows the similar clinical cases extracted by CBR. After the patient's condition is input into the DSS, the CBR base of object cases serves as a corpus to search for optimal treatments for a 40-year-old man with gastric cancer at the postoperative stage IIIa and pyloric obstruction. After separating the new clinical case and the cor-responding health record, CBR computes the similarities of related components by exploring the indexed knowledge. Matching approach is launched to carry out knowledge representation and similar clinical cases selection. CBR seeks to identify clinical cases with important words such as terminologies (e.g. syndrome: acid reflux, belching, vomiting…) and corresponding synonyms, as well as adjectives (e.g. upper abdominal discomfort, occasional postprandial pain…) and verbs (occur, accompany, cause…) that make phrase more accurate.

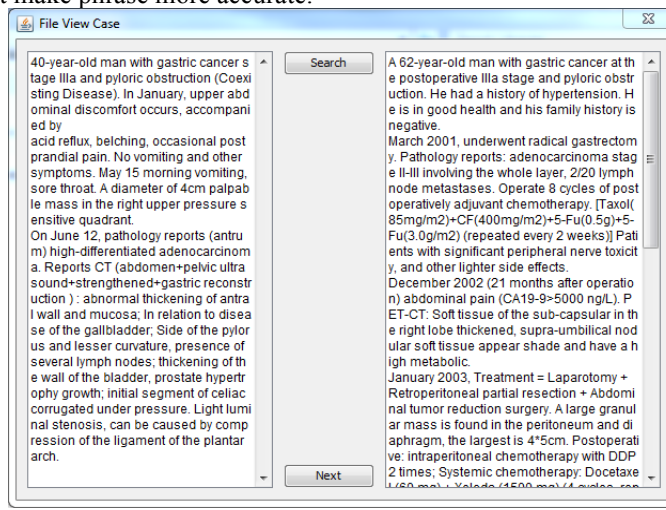

**Figure 6** Extraction of similar clinical cases with CBR

In Fig.6, with the information of current patient such as Δ (gastric cancer stage IIIa) and PB (problem of current case: high-differentiated adenocarcinoma, pyloric obstruction, palpable mass, abnormal thickening of antral wall and mucosa…), DSS works with CBR to search and provide a list of clinical treatment suggestions to patients and physicians.

Concerning the case of Fig.6 top panel (clinical case in need of help), one of the optimal clinical treatment suggestions (Fig.6 bottom panel) is detailed in this article with key elements:
- $1^{st}$ round Θ: underwent radical gastrectomy;
  $2^{nd}$ round Θ: laparotomy+retroperitoneal partial resection+abdominal tumor reduction surgery
  $3^{rd}$ round Θ: chemotherapy: CPT-11(120mg)+5-Fu(300mg);
  $4^{th}$ round Θ: nodular partial excision + side-by-side ascending-ilea colon.
- $1^{st}$ round SΘ: postoperatively adjuvant chemotherapy;
  $2^{nd}$ round SΘ: intraperitoneal chemotherapy with DDP 2;
  $4^{th}$ round SΘ: chemotherapy with DDP +5-FU.

And R (result: death; survival time: 4 years 11 months), as well as some additional information like date and duration. Other extracted similar cases are not detailed for the sake of brevity. Physicians can therefore make clinical decisions by referencing to similar cases provided by DSS.

## 4.2 ROC Graphs for the Performance Evaluation

Based on the positive and negative cases of gastric cancer, we compare the classification effectiveness of the DSS with and without ontology.

The 5-fold cross-validation method was used to randomly divide the experimental dataset into five sub-datasets of the same size. Each was used separately as the test set and the other four as the training set. These five experiments were repeated using the DSS with and without ontology, and the average precision of the classification results was calculated. In a binary classification, a sample is judged to belong to a certain class when its posterior probability of belonging to that class is greater than 0.5. The threshold was therefore set to 0.5 for the precision calculation. The results are presented in Table 1 and Table 2. As shown in these two tables, the average precision of the results increased by 11% (from 73.68% to 84.63%).

**Table 1.** Experimental results on DSS without ontology

| No. of test samples | TP | FP | TN | FN | FP / (FP + TN) | TP/(TP+ FN) | Accuracy |
|---|---|---|---|---|---|---|---|
| 78 | 39 | 17 | 20 | 2 | 0.46 | 0.95 | 75.64% |
| 77 | 38 | 17 | 18 | 3 | 0.48 | 0.93 | 72.72% |
| 77 | 37 | 15 | 19 | 6 | 0.44 | 0.86 | 72.72% |
| 76 | 37 | 16 | 18 | 5 | 0.47 | 0.88 | 72.36% |
| 76 | 37 | 15 | 20 | 4 | 0.43 | 0.90 | 75.00% |
| Average Accuracy | | | | | 73.68% | | |

**Table 2.** Experimental results on DSS with ontology

| No. of test samples | TP | FP | TN | FN | FP / (FP + TN) | TP/(TP+ FN) | Accuracy |
|---|---|---|---|---|---|---|---|
| 78 | 43 | 10 | 23 | 2 | 0.30 | 0.95 | 84.61% |
| 77 | 43 | 7 | 24 | 3 | 0.22 | 0.94 | 87.01% |
| 77 | 42 | 11 | 22 | 2 | 0.33 | 0.95 | 83.11% |
| 76 | 44 | 8 | 21 | 3 | 0.27 | 0.94 | 85.52% |
| 76 | 41 | 12 | 22 | 1 | 0.35 | 0.97 | 82.89% |
| Average Accuracy | | | | | 84.63% | | |

On the basis of the above experimental results, ROC curves were used to evaluate the classification performance of the DSS with and without ontology (Fig. 7). The red and black curves represent the classification performance of DSS with and without ontology, respectively. The diagonal navy gray line shows a random model. The curves climb quickly upward initially, indicating that the model correctly predicted the cases. Given the same test dataset, the classification performance of the DSS with ontology is significantly better than that of the DSS without ontology. Moreover, the AUC of the DSS with ontology is 0.846, which means that in 84.6% of cases, a randomly selected case from the group where the target equals 1 has a higher score than that of a randomly chosen case from the group where the target equals 0. This approach clearly outperforms the DSS without ontology, which achieved an AUC of only 73%.

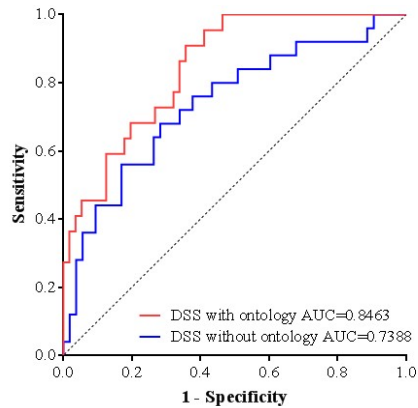

**Figure 7** ROC chart and AUC for classifier evaluations

## 5    Conclusion

This study has developed a DSS for diagnosing and treating cancers. To strengthen the knowledge and interlinking of data, our system is based on the reuse of an existing disease ontology, which improves the reasoning ability of the DSS. The patient's condition is analyzed to estimate the stage of the cancer. Based on previous cases indexed in the CBR database, the search results will be returned to the doctors for use as references in diagnosis.

In our future research, we will adopt the ontology enrichment method to reuse other existing biomedical ontologies, leading to a large domain ontology. In this case, several modules can be added to the DSS. For example, the treatment of cancer patients is conducted in cycles; therefore, a warning function shall be added to the system to remind the medical staff to carry out a new cycle of treatment. When a therapeutic regimen is given by the system according the patient's conditions, an introduction (including pharmaceuticals, price and major efficacy) to the recommended medications [19] will be presented below the window to help doctors use drugs rationally according to the patient's financial situation.

## Acknowledgement

This work was financially supported by the National Natural Science Foundation of China (No.61602013), and the Shenzhen Key Fundamental Research Projects (Grant No. JCYJ20160330095313861, and JCYJ20151030154330711).